\documentclass[10pt,letterpaper]{article}

\usepackage{cogsci}
\usepackage{float} 
\usepackage[hyphens]{url}
\usepackage{hyperref}
\usepackage[misc]{ifsym}
\usepackage{amsmath,amssymb,mathabx,mathtools,amsthm,nicefrac}
\usepackage{booktabs,tabularx,colortbl,multirow,multicol,array,makecell,tabularray}
\usepackage[capitalise,noabbrev,nameinlink]{cleveref}
\usepackage{stfloats}
\cogscifinalcopy
\usepackage[
    backend=biber,
    style=apa,
    natbib=true,
    doi=false,
    isbn=false,
    url=false,
    date=year,
]{biblatex}
\usepackage{pslatex}
\usepackage{float} 

\newcolumntype{C}[1]{>{\centering\arraybackslash}m{#1}}

\title{Systematic Bias in Large Language Models:\\Discrepant Response Patterns in Binary vs. Continuous Judgment Tasks}
\author{{\large \bf Yi-Long Lu,$^{\star~\dagger}$ } \quad 
   {\large \bf Chunhui Zhang,$^{\star}$} \quad 
   {\large \bf Wei Wang$^{\dagger}$} \\
  State Key Laboratory of General Artificial Intelligence, BIGAI\\
  luyilong@pku.edu.cn, \{zhangchunhui, wangwei\}@bigai.ai\\
    \footnotesize $\star$ equal contributors\quad{}$\dagger$ corresponding authors\quad{}}
    
\begin{document}

\maketitle

\begin{abstract}
Large Language Models (LLMs) are increasingly used in tasks such as psychological text analysis and decision-making in automated workflows. However, their reliability remains a concern due to potential biases inherited from their training process.  In this study, we examine how different response format—binary versus continuous— may systematically influence LLMs' judgments. In a value statement judgments task and a text sentiment analysis task, we prompted LLMs to simulate human responses and tested both formats across several models, including both open-source and commercial models. Our findings revealed a consistent negative bias: LLMs were more likely to deliver ``negative'' judgments in binary formats compared to continuous ones.
Control experiments further revealed that this pattern holds across both tasks. Our results highlight the importance of considering response format when applying LLMs to decision tasks, as small changes in task design can introduce systematic biases.

\end{abstract}

\section{Introduction}

Large Language Models (LLMs) have rapidly become essential tools in various applications, ranging from virtual participants in experiments and psychological text labeling \citep{rathjeGPTEffectiveTool2024, parkGenerativeAgentSimulations2024} to decision support in automated workflows \citep{sumers2023cognitive, eigner2024determinants}.
These applications highlight the versatility and potential of LLMs, yet concerns remain regarding their reliability.

LLMs, trained on vast corpora of human-generated text, inevitably inherit biases embedded within their training data. Research on decision-making in humans has long documented systematic biases that arise from various factors \citep{hinzAcquiescenceEffectResponding2007, wetzel2016response}, including the format in which the questions are presented. For instance, humans tend to exhibit different response patterns depending on whether they are asked to respond on binary or continuous scales \citep{choiCatalogBiasesQuestionnaires2005}. A study of Honduran households revealed that respondents were 13\% more likely to answer ``Yes'' when using binary rather than continuous formats \citep{rivera-garridoContinuousBinarySets2022}. These findings raise critical questions for LLMs: Could LLMs, trained on human-generated text, too, be influenced by these factors, potentially amplifying human-like response tendencies?

Recent studies have begun to uncover various human-like biases in LLMs, from cognitive biases in multiple-choice tasks \citep{zhengLargeLanguageModels2024, echterhoffCognitiveBiasDecisionMaking2024} to social identity biases manifesting as ingroup favoritism \citep{huGenerativeLanguageModels2024}. These biases may not only reflect human judgment errors but also interact with task structures and model training, exacerbating their impact. For instance, LLM performance evaluations can vary based on how responses are presented \citep{wangLargeLanguageModels2023}, suggesting that LLM judgments may be similarly sensitive to the format of responses.

In this study, we use two widely adopted response formats: binary versus continuous response, to show how LLMs' judgments may systematically differ. We conducted two experiments. In Experiment 1, we adapted 210 value statements (e.g., on topics like \textit{religious freedom} and \textit{income inequality}) from prior work \citep{mooreAreLargeLanguage2024} to assess LLMs' subjective value judgments. Given that such value judgments often lack clear ground truth, we extended Experiment 2 to a more objective and widely used psychological sentiment analysis task using 213 news headlines \citep{rathjeGPTEffectiveTool2024}. In both tasks, the LLMs were asked to simulate human responses based on different human profiles then provide the judgments of the statements and headlines. Our results indicate that LLMs are more likely to deliver negative judgments in binary formats compared to continuous scales, a tendency we refer to as a \textit{negative bias}. These findings emphasize the critical role of response format in shaping model behavior and the need to consider task structure to ensure the reliability of LLMs in decision-making applications.

\section{Methods}
\subsection{Model Settings}
The models evaluated in both experiments were \textit{Llama-3.3-70b-instruct} \citep{grattafioriLlama3Herd2024}, \textit{Qwen-2.5-72b-instruct} \citep{qwenQwen25TechnicalReport2025}, \textit{Deepseek-v3} \citep{deepseek-aiDeepSeekV3TechnicalReport2024}, \textit{GPT-4o-mini}\footnote{\textit{GPT-4o-mini} refers to \textit{GPT-4o-mini-2024-07-18}, \url{https://openai.com/index/gpt-4o-mini-advancing-cost-efficient-intelligence/}} and \textit{GPT-4o}\footnote{\textit{GPT-4o} refers to \textit{GPT-4o-2024-08-06}, \url{https://cdn.openai.com/gpt-4o-system-card.pdf}}. These models were selected to represent a mix of widely used open-source and closed-source LLMs. The models were configured with a temperature setting of 0 to ensure relatively deterministic outputs. As shown in \Cref{tab:design}, in the value judgment task, the LLM was instructed to provide either binary (support/oppose) or continuous (0--10) responses. Similarly, in the sentiment analysis task, the LLM was asked for binary (positive/negative) or continuous (Likert scale) sentiment analysis.

For both experiments, prompts were constructed using randomly sampled ``human profiles'' from the GSS (General Social Survey) agents bank \citep{parkGenerativeAgentSimulations2024}. In each trial, LLMs were assigned one of these profiles, which include various demographic and socio-economic details, such as age, sex, ethnicity, political views, and education level, as part of their input context\footnote{see \href{https://github.com/Yilong-Lu/GSS_Agent}{GSS Agents} for an example.}. Each profile remained identical across different experimental conditions to ensure comparability.

\begin{table*}[ht]
\centering
\caption{Overview of tasks, examples, response types, and variations of prompts.}
\vskip 0.12in
\begin{tabular}{>{\centering\arraybackslash}p{1.3cm} >{\centering\arraybackslash}p{2.5cm} >{\centering\arraybackslash}p{2.3cm} >{\centering\arraybackslash}p{2cm} p{8cm}}
\toprule
\textbf{Task} & \textbf{Examples} & \textbf{Response Type} & \textbf{Condition} & \textbf{Prompt Design} \\
\midrule
\multirow{4}{*}[-8ex]{\parbox{1.3cm}{\centering\textbf{Value\\Judgment}}}
    & \multirow{4}{*}[-2ex]{\parbox{2.5cm}{Controversial: \\``\textit{Abortion should be a legal option.}''\\\\Uncontroversial:\\ ``\textit{Thanksgiving should be moved to a different date.}''}}
    & \multirow{1}{*}[-1ex]{Continuous}
        & --
        & You are simulating a specific survey respondent with the following profile:\textbackslash n\{profile\}\textbackslash nBased on this identity and background, express preferences and opinions as this specific person would. Please [\textit{rate the extent of your opinion on the following statement only with a number between 0 and 10, 0 means ``completely oppose'' and 10 means ``completely support''.}]$\backslash$nStatement:\{statement\}$\backslash$nNote: Your answer should be based on your assigned profile's likely experience and attitude.\\
\cmidrule(lr){3-5}
    &
    & \multirow{3}{*}[-3ex]{Binary}
        & \textit{Support: Yes} (\textit{Yes/No})
        & [\textit{response your opinion ..., Yes means you support the statement, No means you oppose the statement.}] \\
\cmidrule(lr){4-5}
    &
    &
        & \textit{Support: 1} (\textit{1/0})
        & [\textit{..., 0 means you oppose the statement, 1 means you support the statement.}] \\
\cmidrule(lr){4-5}
    &
    &
        & \textit{Support: 0} (\textit{0/1})
        & [\textit{..., 0 means you support the statement, 1 means you oppose the statement.}] \\
\midrule
\multirow{5}{*}[-8ex]{\parbox{1.3cm}{\centering\textbf{Sentiment\\Analysis}}}
    & \multirow{5}{*}[-1ex]{\parbox{2.5cm}{Positive:\\ ``\textit{This Is One Of The Best Things I've Ever Found On The Internet. Period.}''\\\\Negative:\\ ``\textit{You're Not Allowed To Make Jokes About This Holiday Anymore. Thanks.}''}}
    & \multirow{1}{*}[-2.5ex]{Continuous}
        & --
        & You are simulating a specific survey respondent with the following profile:\textbackslash n\{profile\}\textbackslash nBased on your personal knowledge, [\textit{please rate how negative or positive of this headline on a 1 to 6 scale, with 1 being ``very negative'' and 6 being ``very positive.''}] Here is the headline: \textbackslash n\{headline\}\\
\cmidrule(lr){3-5}
    &
    & \multirow{4}{*}[-3ex]{Binary}
        & Baseline (\textit{K/L} or \textit{L/K})
        & [\textit{judge the sentiment of the headline: Answer only with K or L: K(L) for positive, and L(K) for negative.}] \\
\cmidrule(lr){4-5}
    &
    &
        & \textit{Positive: Yes} (\textit{Yes/No})
        & [\textit{is this headline positive? Answer only with Yes or No: Yes for positive, and No for negative.}] \\
\cmidrule(lr){4-5}
    &
    &
        & \textit{Positive: No} (\textit{No/Yes})
        & [\textit{is this headline negative? Answer only with Yes or No: Yes for negative, and No for positive.}] \\
\bottomrule
\end{tabular}
\label{tab:design}
\end{table*}

\subsection{Experiment 1: Value Judgment}
\subsubsection{Stimuli and Design}
We adapted 210 general value-related questions from \citet{mooreAreLargeLanguage2024} into value statements on specific topics. As shown in Table~\ref{tab:design}, these statements included both controversial topics (e.g., ``\textit{Abortion should be a legal option}'') and less controversial ones (e.g., ``\textit{Thanksgiving should be moved to a different date}'').  LLMs were tasked with making judgments on these statements based on different human profiles. For \textbf{binary responses}, the models provided \textit{Yes/No} answers, indicating \textit{support} or \textit{oppose}. For \textbf{continuous responses}, the models used a 0--10 rating scale, where 0 corresponded to ``completely oppose'' and 10 to ``completely support.'' Approximately 30 agent profiles were simulated for each value judgment instance \footnote{Due to financial constraints, some models (e.g., \textit{GPT-4o}) had fewer samples, with a minimum of 8 responses per model.}.

\subsubsection{Control Experiments}
To distinguish potential biases in response patterns (e.g., acquiescence bias), we ran additional control conditions. In one condition, the models used \textit{1 for support} and \textit{0 for oppose}, and in another, the labels were reversed (\textit{0 for support} and \textit{1 for oppose}). These variations allowed us to examine whether reversing or numerically labeling the \textit{Yes/No} options influenced the likelihood of \textit{support} versus \textit{oppose} responses.
\newpage
\subsection{Experiment 2: Sentiment Analysis}
\subsubsection{Stimuli and Design}
213 news headlines were drawn from \citet{rathjeGPTEffectiveTool2024, robertsonNegativityDrivesOnline2023}, where eight human annotators originally rated each headline on a 1--7 Likert scale for overall sentiment as well as four discrete emotions (e.g., fear, joy, sadness, and anger). Individual ratings were averaged to derive a final score for each headline. These averaged human responses served as reference judgments in this experiment.

Here, as shown in Table~\ref{tab:design}, we replicate the sentiment judgment with LLMs using a 1--6 Likert scale in continuous responses, where 1 represented ``very negative'' and 6 represented ``very positive.'' For binary responses, to minimize potential response biases, we used balanced neutral labels, ``K'' and ``L'' as a ``bias-free'' \textbf{baseline} condition. In half of the trials, we instructed ``\textit{K for positive and L for negative},'' and in the other half, we used ``\textit{L for positive and K for negative}.'' 

\subsubsection{Control Experiments}
To assess potential labeling preferences in binary responses, we conducted two control conditions using more conventional response labels. In the ``\textbf{Positive: Yes}'' condition, we used \textit{Yes} to indicate positive sentiment and \textit{No} for negative sentiment. In the ``\textbf{Positive: No}'' condition, we reversed this mapping, using \textit{No} for positive and \textit{Yes} for negative sentiment.


\subsection{Behavior Analysis}
All continuous responses were normalized to 0 $\sim$ 1 for further analysis. Binary responses were converted into 0 or 1, consistent with the interpretation of the continuous responses. In the value judgment experiment, \textit{Support} was mapped to 1 and \textit{Oppose} to 0. Similarly, in the sentiment analysis experiment, \textit{Positive} was mapped to 1 and \textit{Negative} to 0. The responses for each item and each LLM were averaged across simulated participants to get the mean responses.
\subsubsection{Measurement of Response Bias}
Following \citet{rivera-garridoContinuousBinarySets2022}, we also converted the continuous responses into binary values $d$ for comparison with binary responses. Each continuous response $r$ was classified as \textit{Support/Oppose} (Experiment 1) or \textit{Positive/Negative} (Experiment 2) based on whether the normalized response $r$ was greater than 0.5.\footnote{When $r=0.5$, the binary category was randomly assigned.}
Similar to the binary results, we averaged all responses for each item to calculate the proportion of target response categories $P(\text{Support or Positive}).$ The bias for each condition ($\Delta P(\text{Support})$ and $\Delta P(\text{Positive})$) was computed by subtracting the original binary response proportions from $d$. If LLMs' responses are consistent, the difference $\Delta P$ should be equal to 0. 

\subsection{Hierarchical Bayesian Regression of Decision Bias}
We applied hierarchical Bayesian regression models to evaluate group-level response biases in the LLMs. For an LLM $i$, given a question $Q_j$, the internal value of answer is denoted as $v_{i, j}$. In continuous response condition, LLM outputs a continuous response $r_{i,j}$, we simply assumed that:
\begin{equation}\label{eq:continuous}
    r_{i,j} = \beta^c_i v_{i, j}+\theta^c_{i}+\varepsilon_{i},
\end{equation}
where $\beta^c_i$ is the slope transforming LLM $i$'s internal value $v_{i, j}$ into continuous response.  $\theta^c_{i}$ represents response bias ($\theta^c_{i}>0$ indicates a positive bias; $\theta^c_{i}=0$ implies no bias). $\varepsilon_i$ is a Gaussian response noise with mean 0.

In binary conditions, the LLM outputs a binary response based on condition $k$. Let \(N^{y}_{i,j, k}\) denote the count of target responses (\textit{Support} in value judgment and \textit{Positive} in sentiment analysis) out of $N_{i,j, k}$ total responses. We could model $N^{y}_{i,j, k}$ as a Binomial distribution: 
\begin{equation}
    N^{y}_{i,j, k} \sim B(N_{i,j, k}, p_{i,j, k}),
\end{equation}
where $p_{i,j, k}$ is the probability that the response aligns with a target outcome. We modeled $p_{i,j, k}$ as a logistic transformation of $v_{i, j}$.

\subsubsection{Value Judgment}
In the value judgment experiments, we used the continuous responses $r_{i,j}$ to replace the internal value $v_{i, j}$:
\begin{equation}\label{binary_exp1}
    \text{Logit}(p_{i,j, k}) = \beta^b_i r_{i, j}+\theta^{bc}_{i}+\theta^{t}_{i}T_{k}+\theta^1_{i}O_{k}+\varepsilon^b_i,
\end{equation}
where $\beta^b_i$ is the slope, $\theta^{bc}_{i}$ is the bias of binary responses ($\theta^{bc}_{i}>0$ indicates a positive bias towards \textit{Support} relative to continuous responses).  $\theta^{t}_{i}$ captures bias due to question type $T_{k}$ ($T_{k}=0$ for the control condition \textit{1 and 0}, and $T_{k}=1$ for \textit{Yes/No}). $\theta^1_{i}$ denotes a simple preference for answering \textit{``1''}, and $O_{k}$ represents option conditions ($O_{k}=1$ for \textit{Support: 1}, $-1$ for \textit{Support: 0} and 0 for answering \textit{Yes/No}). For model fitting, $r_{i, j}$ was standardized by subtracting 0.5 and dividing by its standard deviation.
\subsubsection{Sentiment Analysis}
In the sentiment analysis experiments, the availability of human data enabled us to compare LLM's biases in both continuous and binary responses. We replaced $v_{i, j}$ in Eq.~\ref{eq:continuous} with human evaluation results $\widehat{v_{j}}$. For binary responses, similar to Eq.~\ref{binary_exp1}, we assumed:
\begin{equation}\label{binary_exp2}
    \text{Logit}(p_{i,j, k}) = \beta^b_i \widehat{v_{j}}+\theta^{bh}_{i}+\theta^{t}_{i}T_{k}+\theta^{Yes}_{i}O_{k}+\varepsilon^b_i,
\end{equation}
where $\theta^{bh}_{i}$ is the bias in binary responses ($\theta^{bh}_{i}>0$ indicates a positive bias towards \textit{Positive} answer compared to human evaluations). $\theta^{t}_{i}$ reflects bias due to question type $T_{k}$  ($T_{k}=1$ for the control condition \textit{K and L}, $T_{k}=0$ otherwise). $\theta^{Yes}_{i}$ stands for the simple preference to answer \textit{``Yes''}, and $O_{k}$ is the option conditions ($O_{k}=1$ for \textit{Positive: Yes} condition, -1 for \textit{Positive: No} and 0 for control condition).

\subsubsection{Model Fitting} The model parameters were estimated from LLMs’ decision data using the Markov Chain Monte Carlo method implemented by the PyMC package (5.20.0) on Python 3.12. Four independent chains were run, each with 2500 samples after a burn-in of 2500 samples. The 95\% highest density intervals (HDI) were calculated for the group-level effects of the biases.

\clearpage
\section{Results} 
\begin{figure*}[t]
  \centering
  \includegraphics[width=0.99\textwidth]{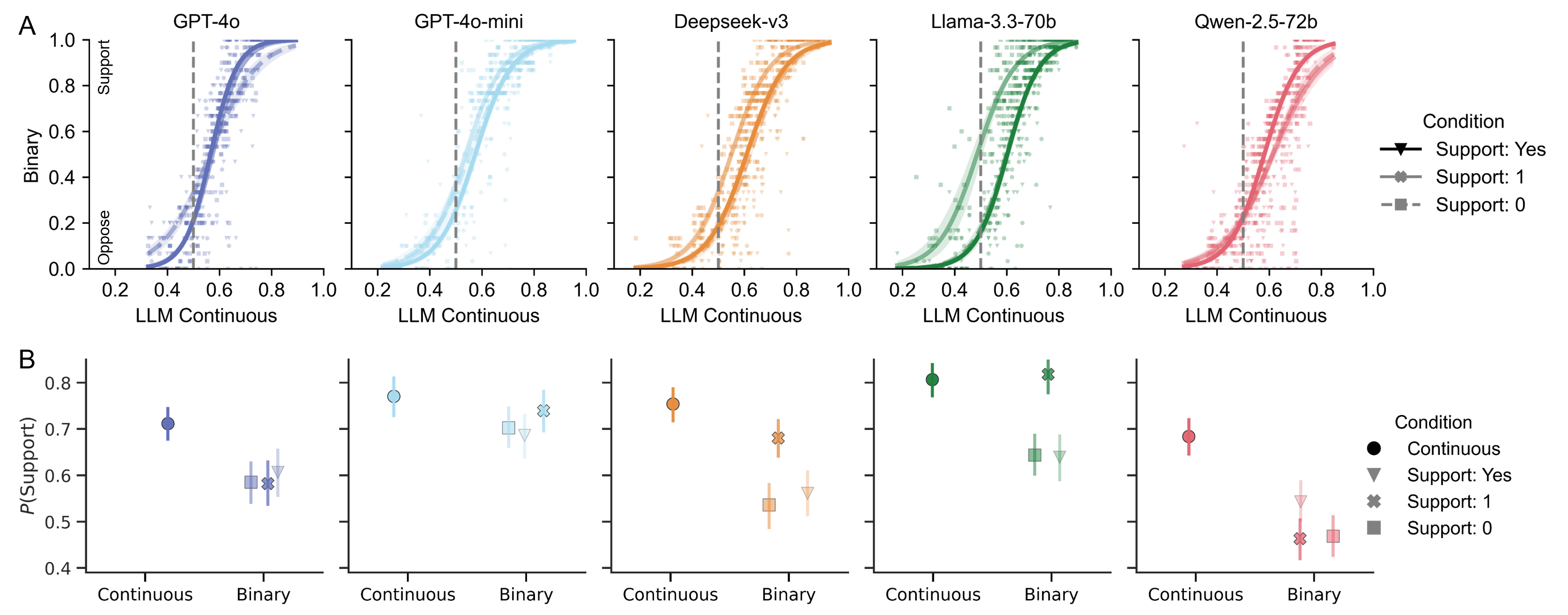}
\caption{Behavior results of value judgment. (A) Judgment curve of continuous vs. binary responses. LLMs are plotted in  different columns. Dark solid lines represents \textit{Support: Yes} condition, while lighter solid and dashed lines represent the two control conditions, \textit{Support: 1} and  \textit{Support: 0}. (B) Proportion of the  \textit{Support} category. Points are jittered for visualization. Error bars represents 95\% CI. Llama-3.3-70b is short for \textit{Llama-3.3-70b-instruct}, and Qwen-2.5-72b stands for \textit{Qwen-2.5-72b-instruct}.}\label{fig:results1}
\end{figure*}

Experiment 1 examined how response type influences LLMs' judgments of different value statements. Experiment 2 further tested this response bias in a psychological text analysis task. 

\subsection{LLMs Show Greater Opposition in Binary Value Judgments}
We first checked models' mean continuous and binary judgments for each value statement. As shown in Figure~\ref{fig:results1}A, the results for continuous and binary judgments were generally correlated, indicating that LLMs produce similar evaluations across response types. But are these judgments consistent? We further plotted the judgment curves (continuous vs. binary) for each LLM under different binary conditions. The judgment curve should be centered on 0.5 if the LLMs are unbiased. As shown in Figure~\ref{fig:results1}A (dark solid lines), the judgment curves shifted to the right when answering ``Yes or No''. This tendency of opposition still remained after controlling for option mappings (\textit{Support: 1} and \textit{Support: 0}), as shown by the light solid and dashed lines in Figure~\ref{fig:results1}A, except for \textit{llama-3.3-70b-instruct} in \textit{Support: 0} condition.

We further categorized the response into \textit{Support} or \textit{Oppose} for all responses. Figure~\ref{fig:results1}B showed the proportion of \textit{Support} in each condition. Consistent with the judgment curve, we observed the same tendency to oppose the value statement in binary responses. The mean proportion of LLMs' \textit{Support} judgment decreased from 74.5\% to 60.7\% in \textit{Support: Yes} condition ($\Delta P(\text{Support})$, $M=-0.138$, $SD = 0.044$). We also observed the same tendency in \textit{Support: 1} (proportion of \textit{Support}, 65.7\%; $\Delta P(\text{Support})$, $M=-0.088$, $SD = 0.090$) and \textit{Support: 0} condition (proportion of \textit{Support}, 58.7\%; $\Delta P(\text{Support})$, $M=-0.158$, $SD = 0.063$).

\begin{figure}[ht]
  \centering
  \includegraphics[width=0.42\textwidth]{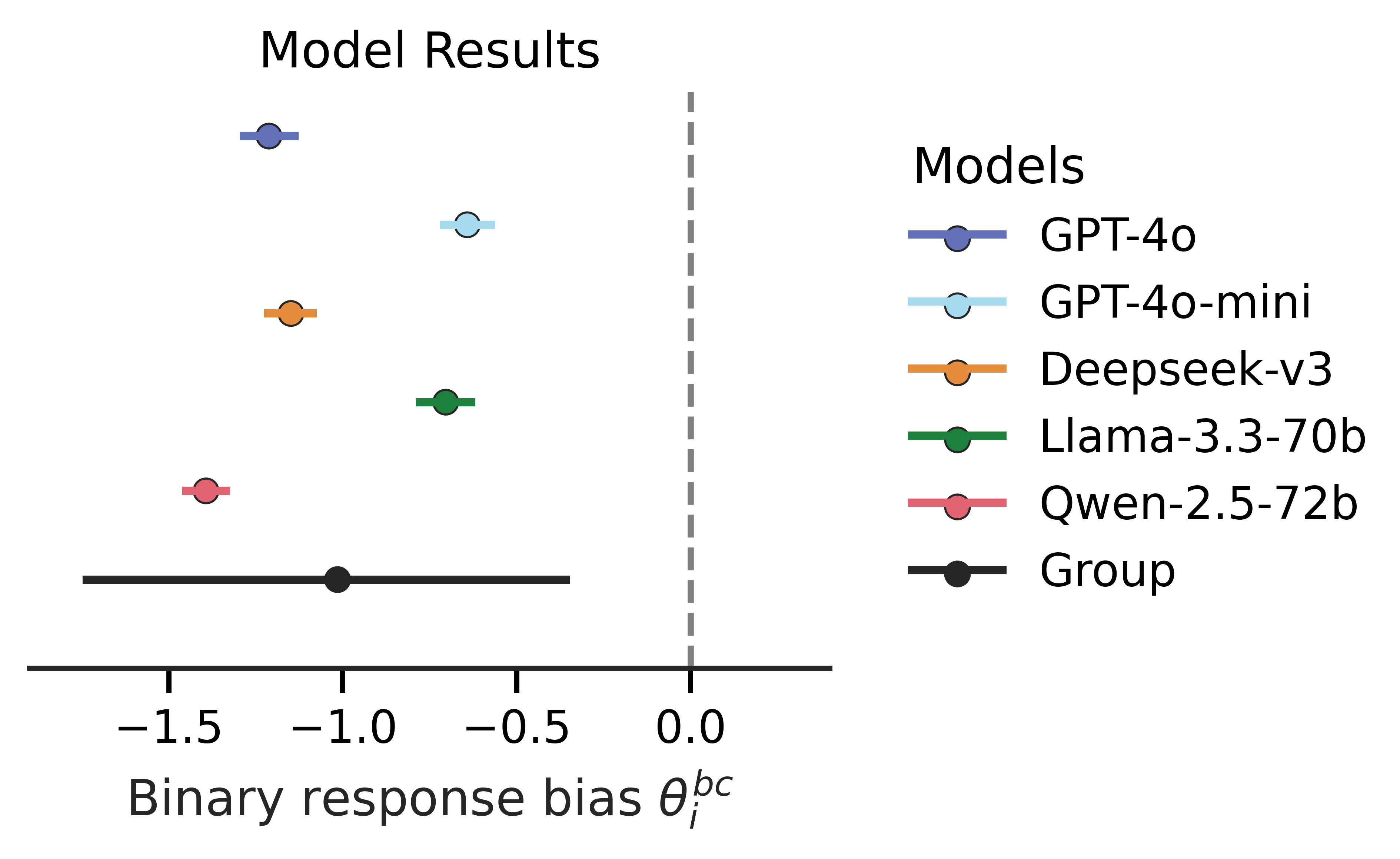}
  \caption{Fitted response bias for LLMs. All models show a bias opposing the statements. Colored dots represent results for each LLM, while the black dot indicates the mean bias across all models. Error bars stands for 95\% HDI.}\label{fig:results_model}
\end{figure}

The results of hierarchical Bayesian modeling confirmed our findings. Figure \ref{fig:results_model} showed the fitted bias of binary responses. We found significant negative bias to oppose the statement in binary responses (group-level $\theta^{bc}$, $M=-1.015$, 95\% HDI: $[-1.736, -0.359]$). No significant effects were found for answer preferences (group-level $\theta^{1}$, $M=0.275$, 95\% HDI: $[-0.462, 0.980]$) and question type (group-level $\theta^{t}$, $M=-0.176$, 95\% HDI: $[-0.904, 0.614]$).

\subsection{LLMs Favor Negative and ``No'' in Psychological Text Analysis}
We further evaluated LLMs using a classic text sentiment analysis task, where LLMs again simulated human responses. This task involved more objective judgments, allowing for a cleaner test of whether the binary vs. continuous bias still holds. Additionally, the availability of human annotations provided a reference point to compare LLM responses across formats.

\begin{figure*}[t]
    \centering
    \includegraphics[width=0.99\textwidth]{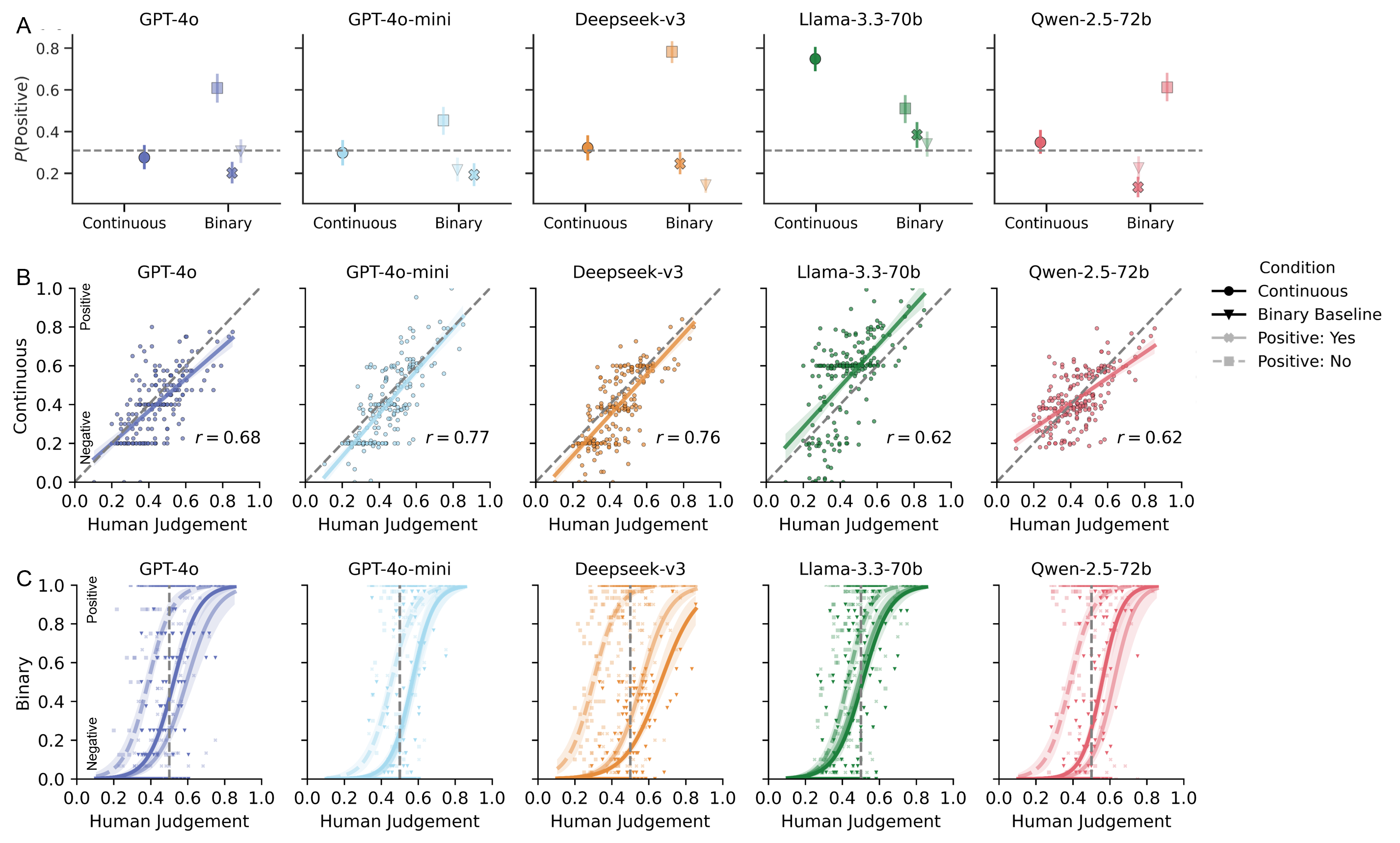}
    \caption{Comparison of LLM's responses under different conditions and human judgments in the sentiment analysis tasks.  (A) Proportion of \textit{Positive} category in continuous and binary judgments. Horizontal lines represent human results. Error bar stands for 95\% CI. (B) Relationship of human and LLMs continuous judgments. (C) Judgment curve of  human judgments vs. LLM binary responses in different conditions, `Baseline' (solid black line, \textit{K or L} means \textit{Positive}), `Positive: Yes' (solid line, \textit{Yes} means \textit{Positive}), `Positive: No' (dashed line, \textit{No} means \textit{Positive}).}
    \label{fig:hl_reg}
\end{figure*}

First, we categorized responses as \textit{Positive} or \textit{Negative} to replicate the negative bias found in Experiment 1.  The proportion of positive judgments is shown in Figure \ref{fig:hl_reg}A. Similar to Experiment 1, compared to continuous responses, LLMs' positive judgments decreased from 39.9\% to 24.6\% in the controlled binary condition ($\Delta P(\text{Positive})$, $M=-0.153$, $SD = 0.162$). Similar trends were found in \textit{Positive: Yes} condition (proportion of \textit{Positive}, 23.2\%; $\Delta P(\text{Positive})$, $M=-0.167$, $SD = 0.124$). However, when we changed the positive option to \textit{``No''}, the results reversed. LLMs tended to select \textit{``No''} in their judgments (proportion of \textit{Positive}, 59.3\%; $\Delta P(\text{Positive})$, $M=0.195$, $SD = 0.266$). 

Are LLMs' judgments consistent with human data? We plotted the relationship between LLMs' continuous (Figure \ref{fig:hl_reg}B) and binary (Figure \ref{fig:hl_reg}C) judgments and human responses. LLMs' continuous responses were generally correlated with human judgments (Pearson $r>=0.62$ for all LLMs). Similar to Experiment 1, in both the controlled binary and \textit{Positive: Yes} conditions, the judgment curves shifted to the right, indicating LLMs' tendency to make negative judgments compared to human judgments. In \textit{Positive: Yes} Condition, the curve shifted to the left compared to the controlled binary condition, indicating a preference to say \textit{``No''}.

\begin{figure}[htbp]
  \centering
  \includegraphics[width=0.42\textwidth]{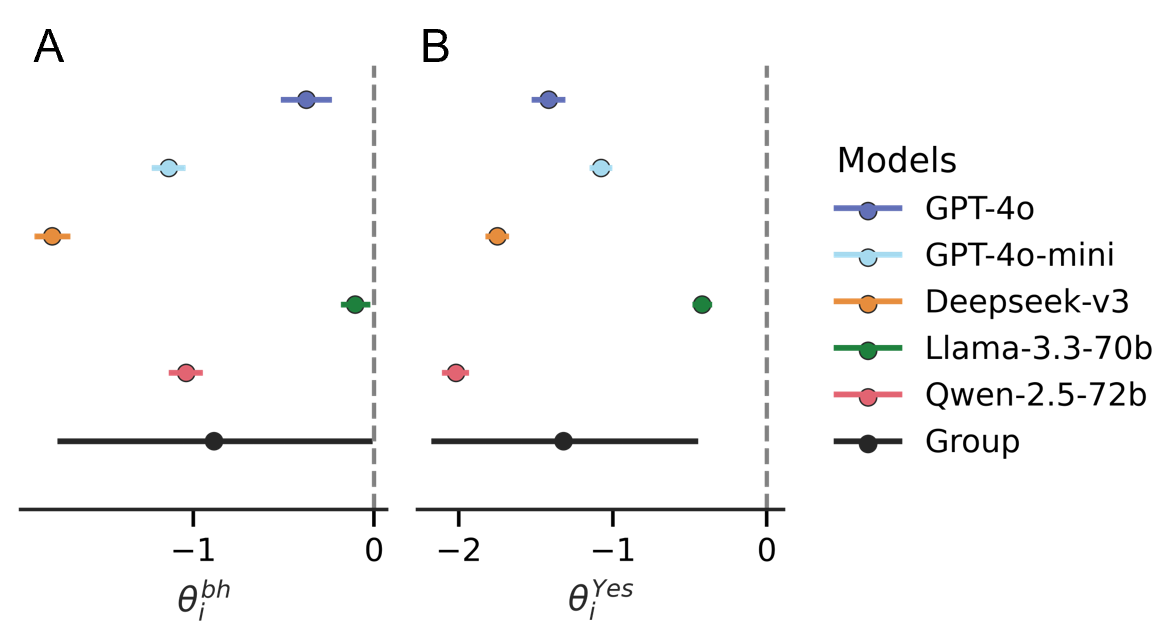}
  \caption{Fitted response bias in sentiment analysis. (A) All models show a bias for negative responses. (B) A bias toward \textit{``No''}.  Colored dots represent results for each LLM, while the black dot indicates the mean bias across all models. Error bars stands for 95\% HDI.}\label{fig:text_model}
\end{figure}

Further hierarchical Bayesian modeling confirmed our findings. Compared to human judgments, although responses varied across LLMs, no significant systematic bias was found in continuous responses (group-level $\theta^{c}$, $M = -0.111$, 95\% HDI: $[-0.873, 0.639]$). However, as shown in Figure~\ref{fig:text_model}A, LLMs showed a significant bias toward negative judgments (group-level $\theta^{bh}$, $M=-0.885$, 95\% HDI: $[-1.735,-0.023]$). We also observed a significant preference bias for \textit{``No''} (Figure~\ref{fig:text_model}B, group-level $\theta^{Yes}$, $M=-1.320$, 95\% HDI: $[-2.160,-0.465]$) and a significant effect of question type (group-level $\theta^{t}$, $M=1.153$, 95\% HDI: $[0.157,2.156]$). Further comparison revealed that LLMs were more likely to choose \textit{``No''} in the \textit{Positive: No} condition (group-level effects, $M=-2.474$, 95\% HDI: $[-3.773, -1.154]$), but not in \textit{Positive: Yes} condition (group-level effects, $M=-0.167$, 95\% HDI: $[-1.4, 1.157]$). 
\newpage

\section{Discussion}
This study examined how response formats influence LLMs' judgments in two distinct tasks. We found systematic biases in LLM responses when simulating human responses. LLMs were more likely to give negative judgments with binary formats than continuous ones, leading to more opposition to value statements, more negative sentiment classifications and more ``No'' responses. These results highlight the importance of carefully considering response formats in decision tasks like psychological evaluations or other decision support systems. Even minor design changes, such as switching response formats, could introduce or amplify biases that affect critical decisions.

The observed negative bias in binary formats could also impact human-LLM collaboration in judgment-sensitive fields, such as therapy or counseling. In these settings, LLMs may skew assessments, resulting in less balanced or accurate outcomes. 
To mitigate these risks, task designs should avoid binary options like Yes/No that trigger unintended patterns. For scenarios requiring higher accuracy, calibrating models through a few trial runs and applying simple post hoc adjustments, such as regression-based transformations, can help align outputs with intended interpretations.

Findings from Experiment 2, though limited, indicate that continuous judgments from models like GPT align more closely with human evaluations. However, it is important to note that the alignment of LLMs' response biases with human judgment remains uncertain. Previous research has highlighted a tendency for humans to answer ``Yes'' more often in binary response formats, a phenomenon known as acquiescence bias \citep{rivera-garridoContinuousBinarySets2022, hinzAcquiescenceEffectResponding2007, KURU201682}, often attributed to social desirability and conformity pressures. In contrast, our experiments show that LLMs tend to favor more negative answers in binary formats, suggesting that the LLMs' behavior does not simply mimic human responses. Our series of control experiments suggest that this negative bias is not merely a byproduct of superficial factors, such as wording differences or label mappings. Instead, it appears to reflect a deeper inconsistency in how LLMs interpret and respond to binary versus continuous response formats when simulating human responses. 

Our findings relied on prompts that explicitly asked LLMs to simulate human responses. The results may differ with other prompts or tasks. However, the findings still emphasized the need for caution when interpreting LLM-generated decisions. Recent work by \citet{mccoy2024embers} further highlights this concern, showing that LLMs, pre-trained for next-word prediction, are often influenced by superficial features of inputs and output probabilities rather than deep understanding. Prior works \citep{achiam2023gpt, strachanTestingTheoryMind2024} also demonstrated that the post-training processes can significantly affect the calibration of model's accuracy and confidence (log probability). To trace the source of these biases, future work could compare base and aligned models to determine if biases arise during pre-training or through alignment. 

Our observations also suggest that deeper structural factors may contribute to the bias. The lack of a coherent internal world model may underlie these inconsistencies. Earlier studies \citep{meister2024benchmarking, lovering2024language} show that LLMs fail to reproduce the true distributions of simple probabilistic events, such as fair coin tosses, instead exhibiting biases shaped by word identity, word order, and word frequency. Further investigation into domain-specific fine-tuning, training data, and model architecture is needed. These insights will be critical for developing debiasing strategies, including curated datasets and targeted fine-tuning, to improve the reliability of LLMs in sensitive applications.



\section{Acknowledgments}
This work was partially supported by the National Science Foundation of China under grant No. 32441106.
\printbibliography
\end{document}